\documentclass[conference]{IEEEtran}
\IEEEoverridecommandlockouts
\usepackage{cite}
\usepackage{amsmath,amssymb,amsfonts}
\usepackage{graphicx}
\usepackage{textcomp}
\usepackage{xcolor}
\usepackage{algorithm}
\usepackage{algpseudocode}
\usepackage{tabularx}
\usepackage{subcaption}
\def\BibTeX{{\rm B\kern-.05em{\sc i\kern-.025em b}\kern-.08em
    T\kern-.1667em\lower.7ex\hbox{E}\kern-.125emX}}
\begin{document}

\title{Adaptive Cluster-Based Synthetic Minority Oversampling Technique for Transportation Mode Choice Prediction with Imbalanced Data\\

}
\author{\IEEEauthorblockN{1\textsuperscript{st} Guang An Ooi}
\IEEEauthorblockA{\textit{Electrical and Computer Engineering} \\
\textit{King Abdullah University of Science and Technology}\\
Thuwal, Saudi Arabia \\
guang.ooi@kaust.edu.sa}
\and
\IEEEauthorblockN{2\textsuperscript{nd} Shehab Ahmed}
\IEEEauthorblockA{\textit{Electrical and Computer Engineering} \\
\textit{King Abdullah University of Science and Technology}\\
Thuwal, Saudi Arabia \\
shehab.ahmed@kaust.edu.sa}
% \and
% \and
% \and
% \and
}
\IEEEpubid{\makebox[\columnwidth]{979-8-3503-6431-6/24/\$31.00~\copyright~2024 IEEE \hfill} \hspace{\columnsep}\makebox[\columnwidth]{ }}
\maketitle

\maketitle

\begin{abstract}
Urban datasets such as citizen transportation modes often contain disproportionately distributed classes, posing significant challenges to the classification of under-represented samples using data-driven models. In the literature, various resampling methods have been developed to create synthetic data for minority classes (oversampling) or remove samples from majority classes (undersampling) to alleviate class imbalance. However, oversampling approaches tend to overgeneralize minor classes that are closely clustered and neglect sparse regions which may contain crucial information. Conversely, undersampling methods potentially remove useful information on certain subgroups. Hence, a resampling approach that takes the inherent distribution of data into consideration is required to ensure appropriate synthetic data creation. This study proposes an adaptive cluster-based synthetic minority oversampling technique. Density-based spatial clustering is applied on minority classes to identify subgroups based on their input features. The classes in each of these subgroups are then oversampled according to the ratio of data points of their local cluster to the largest majority class. When used in conjunction with machine learning models such as random forest and extreme gradient boosting, this oversampling method results in significantly higher F1 scores for the minority classes compared to other resampling techniques. These improved models provide accurate classification of transportation modes.
\end{abstract}


\section{Introduction}
The accurate prediction of individual traffic mode choice is paramount for realistic crowd simulation models in urban areas and effective transportation planning. By analyzing the tendencies of individuals to utilize various modes of traffic such as walking, cycling, public transportation, and private vehicles, accurate traffic demand forecast models can be created to alleviate traffic congestion, optimize the allocation of transportation resources, and foster the adoption of sustainable transportation modes. However, traffic demand data is inherently imbalanced, which poses significant challenges to data-driven models, especially for minority classes in the datasets.

Class imbalance is a result of the skewed distribution of travel modes in a set of data, with certain modes exhibiting significantly higher prevalence than others. This phenomenon is particularly severe in urban environments, where the usage of public transportation and private vehicles are predominant, while slower modes of travel such as walking and cycling are uncommon, especially during work hours. This imbalance can severely bias machine learning algorithms, which are commonly employed for classification problems such as mode choice prediction. When trained on imbalanced data, such algorithms tend to converge on local optima which are easily reached through gradient descent by optimizing their predictions for dominant classes. Such biased models tend to erroneously classify minority classes, leading to the overestimation of The consequences of such bias can be detrimental to transportation planning efforts, as underestimating the demand for sustainable modes like cycling or public transport can impede initiatives aimed at promoting these environmentally friendly alternatives.

Traditional methodologies for travel mode choice prediction, such as discrete choice models (DCM) \cite{so1995multinomial,hensher2018applied}, often suffer from erroneous parameter estimations caused by class imbalance. While DCMs interpretable decision-making processes, their underlying assumption of the independence of irrelevant alternatives (IIA) and homogeneous preferences are offset by imbalanced data. As such, minority classes are often overlooked by the biased parameters of the models.

In recent years, machine learning algorithms such as support vector machines (SVM) \cite{hearst1998support}, extreme gradient boosting (XGBoost) \cite{chen2015xgboost}, and deep neural networks (DNN) \cite{cichy2019deep} have demonstrated significant improvements in feature recognition and learning capabilities, especially for complex datasets with high dimensionalities and sparse distributions. However, these approaches are also susceptible to class imbalance, which offsets their predictions by introducing biases in their parameters \cite{chen2023travel}. 

In this study, an adaptive cluster-based synthetic minority oversampling technique (AC-SMOTE) and its application to imbalanced data in transportation mode choice prediction is proposed. The rest of the paper is structured as follows: Section II presents a detailed literature review of existing resampling techniques and their limitations, establishing the need for the proposed method. Section III outlines the methodology, including dataset selection, model evaluation, and the development of AC-SMOTE. Section IV discusses the results and analyzes the performance of the proposed methodology and benchmarks against contemporary methods in the literature. Finally, Section V concludes the paper by summarizing the findings and discussing the implications of this research for future studies in transportation planning and data-driven modeling. 

\section{Literature Review}

To address the challenges introduced by class imbalance, various oversampling and undersampling techniques have been developed in the literature. Oversampling methods, such as the Synthetic Minority Oversampling Technique (SMOTE), seek to balance the class distribution by generating synthetic samples of the minority classes. Conversely, undersampling techniques aim to achieve balance by reducing the number of samples in the majority class. While these techniques have shown promise in certain contexts, their effectiveness can be contingent upon the specific characteristics of the dataset and the choice of prediction model. Random oversampling and undersampling (ROS and RUS) are two of the most commonly used techniques in managing imbalanced data due to their effectiveness. ROS randomly selects samples in the minor classes within a dataset and creates duplicates to increase their population, while RUS randomly removes samples from major classes to match the amount of minor classes. For ROS, more sophisticated approaches have been developed to create synthetic samples that are similar to real data in most features but possess sufficient difference such that overfitting can be avoided. For example, the synthetic minority oversampling technique (SMOTE) generates artificial samples in the minor classes by calculating the weighted average between a real sample and one of its k-nearest neighbors. However, datasets created by SMOTE tend to be offset by artificial distributions introduced by the synthetic data, which impedes the ability of the machine learning model to learn the actual patterns in real data. SMOTE Nominal and Continuous (SMOTE-NC) is an extended version of SMOTE that allows the algorithm to process nominal data types by selecting the most frequent category among the nearest neighbors and assigning it to the synthetic sample. 

Instead of interpolating between existing minority class samples to generate synthetic data, Borderline-SMOTE is proposed in \cite{han2005borderline} as an oversampling method for imbalanced datasets by specifically generating synthetic samples near the boundary of decisions between minority and majority classes. Borderline samples are identified according to the k-nearest neighbors of each minority sample. This approach is based on the fact that minority class samples near these boundaries have a higher likelihood to be misclassified. Another oversampling method, SVM-SMOTE, is proposed in \cite{bagui2023resampling}, using the SVM classifier instead of k-nearest neighbors to identify minority samples near decision boundary regions. 

On the other hand, the adaptive synthetic sampling approach (ADASYN) evaluates the difficulty in classifying the samples in minor classes through k-nearest neighbors and generates synthetic data for such samples. The primary drawback of this approach is that it is sensitive to outliers and noisy data, which would be categorized as difficult samples to classify by the k-nearest neighbors approach. Additionally, this method tends of bias the machine learning models towards difficult samples, which hinders their ability to correctly classify more commonly seen samples.  

% Conversely, the neighborhood cleaning rule (NCR) is an advanced undersampling approach to remove noisy and outlying samples from a dataset. For an arbitrary number of neighbors around each sample, the NCR algorithm removes any samples that are not classified as the major class in the neighborhood. This process iterates until a certain criterion is achieved. Since this approach removes all minor classes in the given neighborhood, it tends to eliminate samples that represent useful but under-represented information, which causes the training dataset to be incomplete. It also requires careful tuning and is computationally expensive since it iterates through each sample in the dataset. 

In \cite{hagenauer2017comparative}, the class distribution in the Dutch national travel survey (NTS) data \cite{smit2017innovation} is mitigated through the integration of additional data and random oversampling. The performance of seven classification models such as SVM and MNL are evaluated. However, while it is stated that the dataset contains imbalanced classes, the study did not compare the accuracy of the models. \cite{pirra2017tour} proposed a modified SVM method that assigns different weights to different classes in the decision function, demonstrating improved performance compared to the plain SVM. Similarly, \cite{qian2022pointnext} introduced a kernel scaling adjustment in SVM to enhance the accuracy of minority class classification. However, both of these methods are specific to SVM and may not generalize to other machine learning models.

% \cite{kim2022bridging} employed a class-specific weighting scheme in which each instance is assigned weights inversely proportional to the frequency distribution of classes. While this approach acknowledges the importance of minority classes, it treats all instances within a class as equally important, potentially overlooking the diversity within each class.

A recent study \cite{chen2023travel} provides valuable insights into the impact of imbalanced classes in datasets and compares the performance of multiple resampling approaches on the London Passenger Mode Choice (LPMC) dataset. It is shown that, despite improvements achieved by machine learning-based algorithms, minority classes in large datasets are still predicted with significantly less precision and accuracy, which compromises transportation planning and resource allocation.

\section{Methodology}
This section investigates the application of data processing and resampling methods in conjunction with machine learning models such as random forest, XGBoost, and DNNs on imbalanced data. Furthermore, AC-SMOTE is developed to enhance the accuracy of predicting minor traffic modes while maintaining performance on major modes. The combinations of data resampling and machine learning models are evaluated using both general and class-specific metrics.

\subsection{Dataset Selection}
The LPMC dataset is used in this study. This dataset consists of 81,086 trips made by 31,954 individuals from April 2012 to March 2015, and is investigated in recent literature \cite{chen2023travel} due to its inclusiveness and amount of features. The travel modes are classified into Walking, Cycling, Public Transport, and Driving, which occupy 17.80\%, 3.27\%, 36.10\%, and 42.83\% of the dataset, respectively. Cycling samples occupy a significantly smaller portion of the dataset than walking, the second least common mode of travel, taking up 17.80\%. On the contrary, the most common travel mode is driving, occupying 42.83\% of the collected samples, which introduces significant challenges in predicting the minor modes of travel. 
% \begin{table}[!t]
%     \renewcommand{\arraystretch}{1.3}
%     \caption{Travel Mode Distribution}\label{tab:tab1}
%     \centering
%     \resizebox{7.5cm}{!}{
%         \begin{tabular}{l c c}
%             \hline\hline \\[-3mm]
%             \multicolumn{1}{c}{Travel Mode} & \multicolumn{1}{c}{Total Trips} & \multicolumn{1}{c}{Percentage (\%)} \\[1.4ex]\hline
%             Walking & 14268 & 17.80 \\
%             Cycling & 2405 & 3.27 \\
%             Public Transport & 28605 & 36.10 \\
%             Driving & 35808 & 42.83 \\
%             \hline\hline
%         \end{tabular}
%     }
% \end{table}

There are 35 unique features for each individual trip in the original LPMC dataset, such as the purpose of the trip, the type of fuel consumed, and travel month and date. To ensure fair comparison, 14 features in addition to the travel mode are selected in accordance with \cite{chen2023travel} and \cite{wang2020deep}. The durations and costs to reach the destination of each sample are approximated using online resources. All continuous features are standardized as $z = (x - \mu)/(\sigma)$, where \(z\) is the standardized value, \(x\) is the original feature value, \(\mu\) is the mean value of the feature, and \(\sigma\) is the standard deviation of the feature.
% \begin{table}[!t]
%     \renewcommand{\arraystretch}{1.3}
%     \caption{Features Description}\label{tab:tab2}
%     \centering
%     \begin{tabularx}{7.5cm}{l >{\centering\arraybackslash}X}
%         \hline\hline \\[-3mm]
%         \multicolumn{1}{c}{Feature} & \multicolumn{1}{c}{Type} \\[1.4ex]\hline
%         Car ownership & Nominal \\
%         Gender & Nominal \\
%         Driving license & Nominal \\
%         Number of public transport interchanges & Nominal \\
%         Age & Continuous \\
%         Distance & Continuous \\
%         Walking duration & Continuous \\
%         Cycling duration & Continuous \\
%         Duration to access public transport & Continuous \\
%         Total public transport duration & Continuous \\
%         Public transport interchange duration & Continuous \\
%         Driving duration & Continuous \\
%         Cost of public transport & Continuous \\
%         Cost of driving & Continuous \\
%         Travel mode & Nominal \\
%         \hline\hline
%     \end{tabularx}
% \end{table}

Adopting the same ratio as \cite{chen2023travel}, the training dataset consists of samples that are randomly selected from 67.54\% of the dataset while the test dataset consists of the remaining 32.46\%. However, to ensure that the evaluations are generalizable and unbiased, each performance analysis in the following sections is conducted five times, each round using new randomly selected data in the same ratio to provide the average value of the metrics. 

\subsection{Model Evaluation}
Since this study aims to improve predictive performance on minor classes, a class-specific metric is required to evaluate the models. The F1 score, which is the harmonic mean of precision and recall, is formulated as:
\begin{equation}
\begin{aligned}
\text{F1 Score} &= \frac{2 \times \text{Precision} \times \text{Recall}}{\text{Precision} + \text{Recall}}
\end{aligned}
\end{equation}

% \begin{equation}
% \text{Precision} = \frac{\text{TP}}{\text{TP} + \text{FP}}
% \end{equation}

% \begin{equation}
% \text{Recall} = \frac{\text{TP}}{\text{TP} + \text{FN}}
% \end{equation}
where $\text{precision} = \frac{\text{TP}}{\text{TP} + \text{FP}}$ and $\text{Recall} = \frac{\text{TP}}{\text{TP} + \text{FN}}$, and TP, FP, and FN are the true positives, false positives, and false negatives for a specific class. The F1 score is an effective method to evaluate minority class predictions as it considers both precision and recall, which are the ratios of correctly predicted samples within a class to all samples predicted as the class and to all true samples of the class, respectively.

\subsection{Data Resampling}
ROS, RUS, SMOTENC, and ADASYN, which are four of the data resampling methods in \cite{chen2023travel} that provided the highest F1 score for the minor class (cycling), are included to serve as benchmarks against the AC-SMOTE method developed in this study. Data processed by each resampling method are used to train random forest and XGBoost algorithms, as well as fully connected neural networks (FCNNs) to create predictive models. 

The AC-SMOTE algorithm is detailed in Algorithm~\ref{alg:AC-SMOTE}. In comparison to traditional SMOTE, which generates synthetic data based on the k-nearest neighbors of minority class samples, the AC-SMOTE algorithm incorporates a density-based clustering method. This process segments the minority classes into smaller and more homogeneous subgroups that are not categorized by the default labels of the dataset, improving the quality of synthetic data generation and avoiding oversampling outliers. Classes that exceed an arbitrary amount or proportion of samples within the dataset are categorized as the majority, while all remaining classes are labeled as the minority.
\begin{algorithm}
\caption{AC-SMOTE}
\label{alg:AC-SMOTE}
\begin{algorithmic}[1]
\Require $X_{\text{train}}$, $y_{\text{train}}$  \Comment{Training data and labels}
\Require $n_{\text{clusters}}$ \Comment{Number of clusters for K-means}

\State \textbf{Split:} Divide data into minority and majority classes
\State $X_{\text{minority}} \gets$ samples with minority label
\State $y_{\text{minority}} \gets$ corresponding labels

\State \textbf{Cluster:} Apply K-means to $X_{\text{minority}}$
\State $cluster\_labels \gets \text{DBScan}(X_{\text{minority}}, n_{\text{clusters}})$

\State $majority\_size \gets$ number of majority class samples

\For{each $cluster\_label$ in $cluster\_labels$}
    \State $X_{\text{cluster}} \gets$ samples in current cluster
    \State $y_{\text{cluster}} \gets$ labels in current cluster

    \State $cluster\_size \gets$ size of $X_{\text{cluster}}$
    \State $target\_size \gets \max(majority\_size, cluster\_size)$

    \State $(X_{\text{over}}, y_{\text{over}}) \gets \text{SMOTE}(X_{\text{cluster}}, y_{\text{cluster}}, target\_size)$

    \State Append $X_{\text{over}}$ to $X_{\text{resampled}}$
    \State Append $y_{\text{over}}$ to $y_{\text{resampled}}$
\EndFor

\State $X_{\text{train\_balanced}} \gets$ concatenate majority samples and $X_{\text{resampled}}$
\State $y_{\text{train\_balanced}} \gets$ concatenate majority labels and $y_{\text{resampled}}$

\Return $X_{\text{train\_balanced}}, y_{\text{train\_balanced}}$
\end{algorithmic}
\end{algorithm}

In this approach, subgroups are identified via density-based spatial clustering of applications with noise (DBSCAN). DBSCAN is selected over K-means clustering as it utilizes the density of the samples rather than their centroids to perform clustering, which allows noisy and outlying data to be easily identified in low-density regions. This approach is well-suited for classes with sparsely distributed samples. DBSCAN also does not assume spherical and even distribution of samples, allowing it to find arbitrarily-shaped clusters. The clustering operation of DBSCAN can be tuned using parameters such as the maximum distance between points in the same cluster $\epsilon$ and the minimum number of data points in the same cluster $N$. 

\begin{figure}[!t] 
    \centering
    \includegraphics[width=0.37\textwidth]{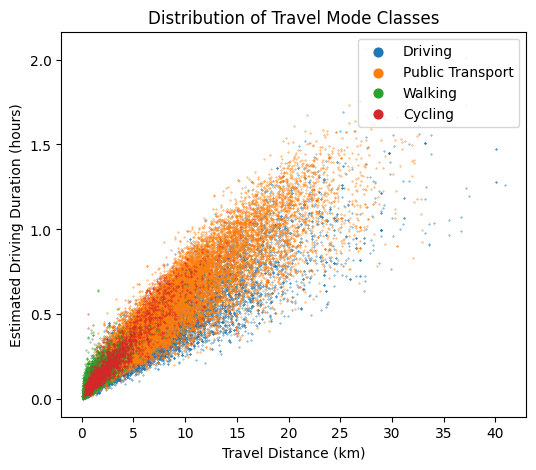}
    \caption{The distribution of the travel modes in the dataset.}
    \label{fig:org_classes}
\end{figure}

\begin{figure}[!t] 
    \centering
    \includegraphics[width=0.37\textwidth]{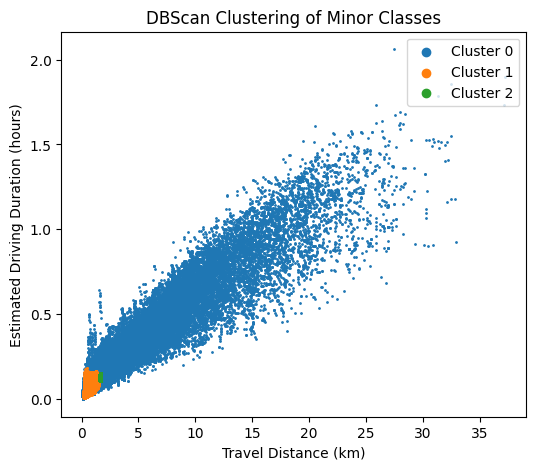}
    \caption{DBScan clustering of "Public Transport," "Walking," and "Cycling" classes.}
    \label{fig:dbscan}
\end{figure}

Fig.~\ref{fig:org_classes} shows the distribution of the four travel modes, while Fig.~\ref{fig:dbscan} shows an example of clustering with $\epsilon = 33$ and $N = 99$ plotted with the estimated driving duration and travel distance as the y- and x-axes, respectively. It is found through grid search that this set of $\epsilon$ and $N$ values resulted in the highest average F1 score in the prediction of all minor classes. Cluster 0, which occupies a great portion of the minority classes, represents all noisy data that failed to be groups within an actual cluster. This shows the sparse distribution as well as high variability of these data, which renders it difficult for machine learning models to effectively learn to identify these minor classes. Furthermore, Clusters 1 and 2 (orange and green points, respectively) are two distinctly smaller groups that occupy the bottom left part of the figure. Such clusters are created using features in the dataset without relying on the labels, which provides insights into the diversities within the minor classes and avoids undesired generalization in oversampling. 

\begin{figure*}[!htb]
    \centering
    \begin{subfigure}{0.3\textwidth}
        \centering
        \includegraphics[width=\textwidth]{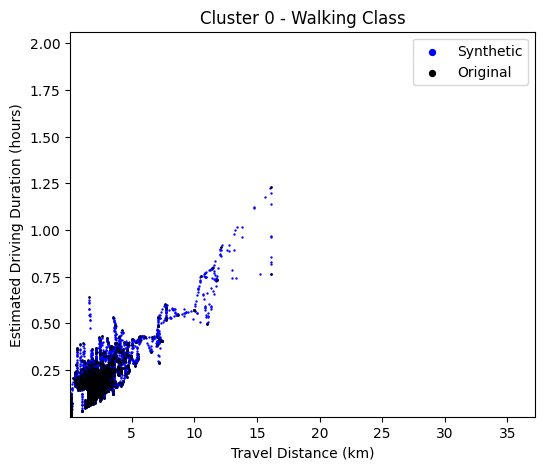}
        % \caption{}
        \label{fig:subfig0}
    \end{subfigure}
    \begin{subfigure}{0.3\textwidth}
        \centering
        \includegraphics[width=\textwidth]{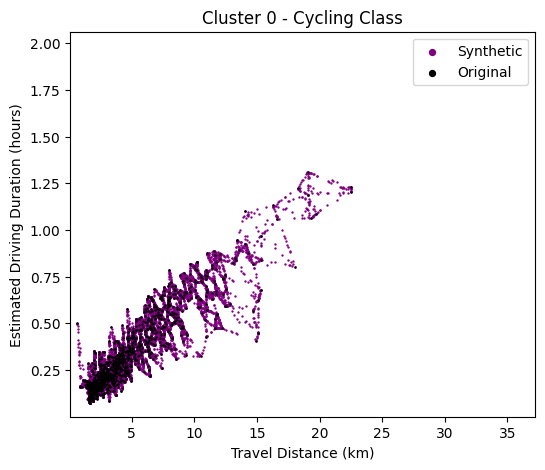}
        % \caption{}
        \label{fig:subfig1}
    \end{subfigure}
    \begin{subfigure}{0.3\textwidth}
        \centering
        \includegraphics[width=\textwidth]{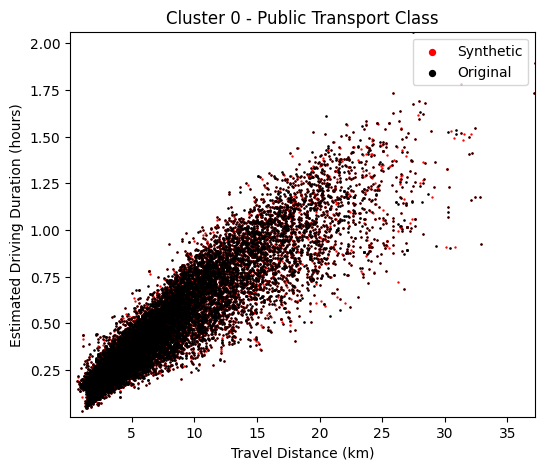}
        % \caption{}
        \label{fig:subfig2}
    \end{subfigure}
    \begin{subfigure}{0.3\textwidth}
        \centering
        \includegraphics[width=\textwidth]{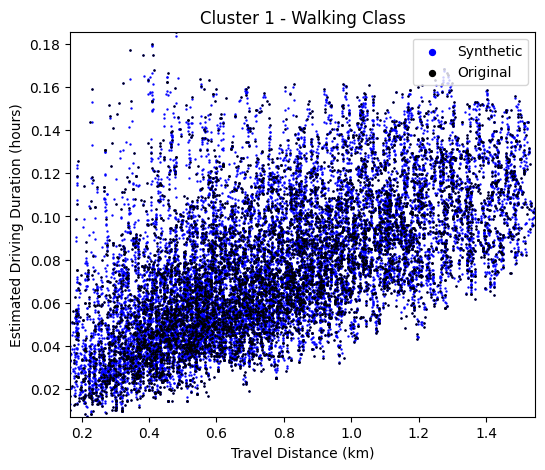}
        % \caption{}
        \label{fig:subfig3}
    \end{subfigure}
    \begin{subfigure}{0.3\textwidth}
        \centering
        \includegraphics[width=\textwidth]{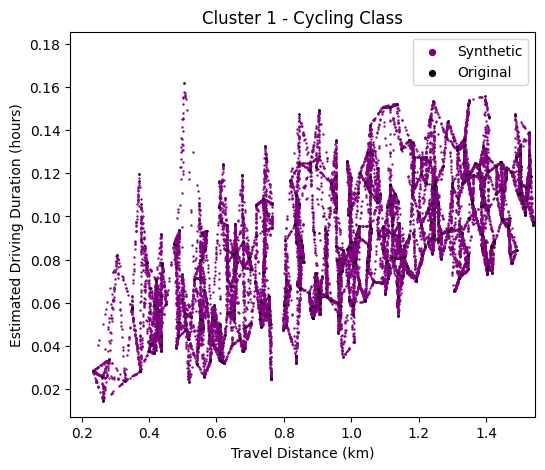}
        % \caption{}
        \label{fig:subfig4}
    \end{subfigure}
    \begin{subfigure}{0.3\textwidth}
        \centering
        \includegraphics[width=\textwidth]{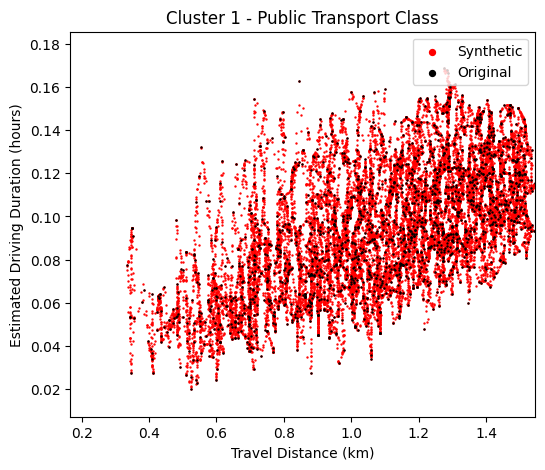}
        % \caption{}
        \label{fig:subfig5}
    \end{subfigure}
    
    \begin{subfigure}{0.3\textwidth}
        \centering
        \includegraphics[width=\textwidth]{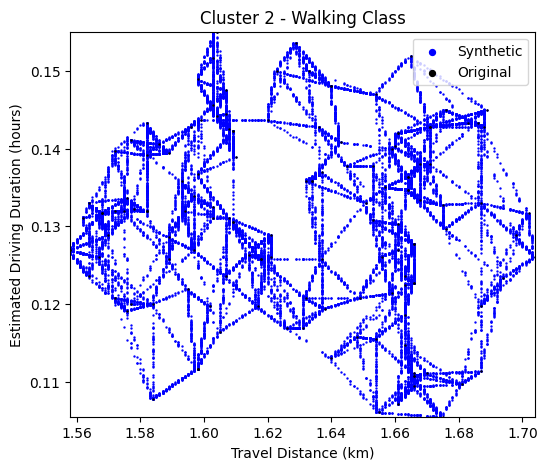}
        % \caption{}
        \label{fig:subfig6}
    \end{subfigure}
    \begin{subfigure}{0.3\textwidth}
        \centering
        \includegraphics[width=\textwidth]{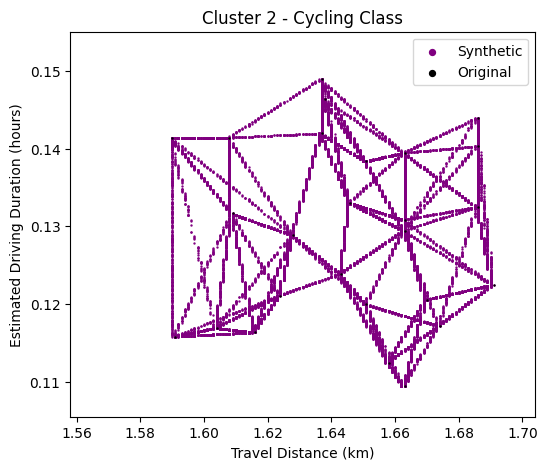}
        % \caption{}
        \label{fig:subfig7}
    \end{subfigure}
    \begin{subfigure}{0.3\textwidth}
        \centering
        \includegraphics[width=\textwidth]{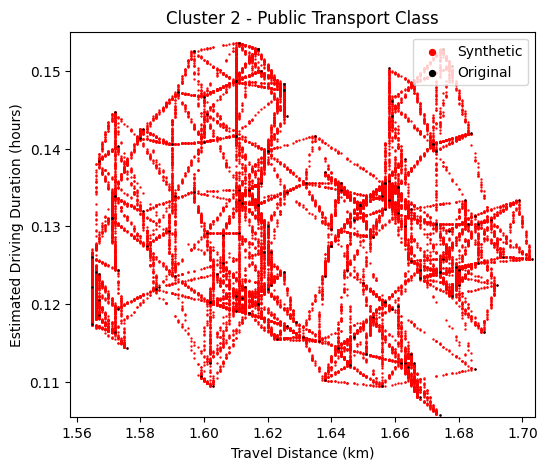}
        % \caption{}
        \label{fig:subfig8}
    \end{subfigure}
    \caption{Synthetic data creation for the minority classes.}
    \label{fig:3x3}
\end{figure*}

SMOTE is then applied on samples in each of these clusters that belong to the same class by iterating through the clusters. The amount of synthetic data to be generated for each class is determined by selecting the larger value between the size of the cluster and the size of the largest major class. The scatter plots in Fig.~\ref{fig:3x3} show the distribution of the three minor classes in Clusters 0, 1, and 2 in black and the synthetic data generated for each class in blue, purple, and red. Walking, cycling, and public transport classes are clearly visible in significant proportions in all clusters, which highlights the importance of clustering the samples according to their features prior to oversampling. 

\section{Results and Discussion}
The original dataset, along with datasets processed using ROS, RUS, SMOTENC, ADASYN, and AC-SMOTE, are separately used to train random forest, XGBoost, and FCNN models. The parameters of the XGBoost model are optimized through grid search, as detailed in Table~\ref{tab:xgboost}. Furthermore, to address the imbalance of classes in the training data, the weight of each class is calculated as \(W_i = N_max/N_i\), where \(W_i\) is the weight of the \(i^{th}\) class, \(N_max\) is the highest number of samples per class, and \(N_i\) is the number of samples of the \(i^{th}\) class.
\begin{table}[!t]
    \renewcommand{\arraystretch}{1.3}
    \caption{XGBoost Hyperparameter Optimization}\label{tab:xgboost}
    \centering
    \begin{tabularx}{7.5cm}{X c c}
        \hline\hline \\[-3mm]
        Parameter & Grid Range & Optimal Value \\[1.4ex]\hline
        Learning Rate & 0.1:0.1:1.0 & 0.8 \\
        Maximum Depth & 2:2:20 & 6 \\
        Minimum Child Weight & 1:1:10 & 1 \\
        Subsampling ratio & 0.1:0.1:1.0 & 0.8 \\
        Feature subsampling ratio & 0.1:0.1:1.0 & 0.8 \\
        \hline\hline
    \end{tabularx}
\end{table}

The hyperparameters of the FCNN is also optimized via grid search similarly to XGBoost, as shown in Table~\ref{tab:fcnn}. 
\begin{table}[!t]
    \renewcommand{\arraystretch}{1.3}
    \caption{FCNN Hyperparameter Optimization}\label{tab:fcnn}
    \centering
    \begin{tabularx}{7.5cm}{X c c}
        \hline\hline \\[-3mm]
        Parameter & Grid Range & Optimal Value \\[1.4ex]\hline
        Learning Rate & 0.001:0.005:0.1 & 0.061 \\
        Number of layers & 1:1:10 & 2 \\
        Neurons per layer & $2e^N, N = 4:1:10$ & $2e^7$ \\
        Dropout rate & 0.1:0.1:0.9 & 0.2 \\
        \hline\hline
    \end{tabularx}
\end{table}

The F1 score of each of these combinations in predicting the individual travel mode classes in the test dataset are listed in Tables IV, V, VI, and VII. The distribution of the results of combining the former five types of data with random forest and XGBoost is similar to \cite{chen2023travel}. In most instances, the Driving class is predicted with the highest F1 score due to its large amount of samples in the dataset. A slight increase in performance is observed in the best F1 scores for Walking, Public Transport, and Driving as compared to \cite{chen2023travel}. For these three classes, the original and AC-SMOTE-augmented data both achieved the highest F1 score when combined with XGBoost. As observed in \cite{chen2023travel}, the F1 scores of all combinations for the Cycling class is significantly lower than other classes due to its low sample count. However, the AC-SMOTE data produced the highest F1 score when combined with the random forest algorithm. This combination also resulted in relatively high F1 scores for other classes. ADASYN also produced relatively good F1 scores when combined with the random forest and XGBoost models. 
\begin{table*}[!t]
    \renewcommand{\arraystretch}{1.3}
    \centering    
    \begin{minipage}[b]{0.48\textwidth}
        \centering
        \caption{Walking Class Prediction F1 Scores}
        \begin{tabular}{l c c c c}
            \hline\hline
            \multicolumn{1}{c}{Method} & \multicolumn{1}{c}{Random Forest} & \multicolumn{1}{c}{XGBoost} & \multicolumn{1}{c}{FCNN} & \multicolumn{1}{c}{\begin{tabular}[c]{@{}c@{}}Hierarchical \\ Clustering\end{tabular}} \\ \hline
            Original & \textbf{0.74} & \textbf{0.74} & 0.67 & 0.73\\
            ROS & 0.63 & 0.69 & 0.66 & 0.66 \\
            RUS & 0.66 & 0.64 & 0.58 & 0.66 \\
            SMOTE-NC & 0.73 & 0.69 & 0.64 & 0.73 \\
            ADASYN & 0.72 & 0.73 & 0.58 & 0.73 \\
            Borderline-SMOTE & 0.71 & 0.68 & 0.72 & 0.73 \\
            SVM-SMOTE & 0.72 & 0.68 & 0.69 & 0.72 \\
            AC-SMOTE & 0.73 & \textbf{0.74} & 0.66 & 0.73 \\
            \hline\hline
        \end{tabular}
    \end{minipage}%
    \hfill
    \begin{minipage}[b]{0.48\textwidth}
        \centering
        \caption{Cycling Class Prediction F1 Scores}
        \begin{tabular}{l c c c c}
            \hline\hline
            \multicolumn{1}{c}{Method} & \multicolumn{1}{c}{Random Forest} & \multicolumn{1}{c}{XGBoost} & \multicolumn{1}{c}{FCNN} & \multicolumn{1}{c}{\begin{tabular}[c]{@{}c@{}}Hierarchical \\ Clustering\end{tabular}} \\ \hline
            Original & 0.09 & 0.15 & 0.00 & 0.09\\
            ROS & 0.07 & 0.22 & 0.13 & 0.20 \\
            RUS & 0.08 & 0.18 & 0.00 & 0.20 \\
            SMOTE-NC & 0.19 & 0.18 & 0.12 & 0.08 \\
            ADASYN & 0.22 & 0.23 & 0.10 & 0.08 \\
            Borderline-SMOTE & 0.22 & 0.18 & 0.21 & 0.08 \\
            SVM-SMOTE & 0.22 & 0.15 & 0.10 & 0.03 \\
            AC-SMOTE & \textbf{0.27} & 0.23 & 0.14 & 0.08\\
            \hline\hline
        \end{tabular}
    \end{minipage}
    
    \vspace{0.3cm}
    
    \begin{minipage}[b]{0.48\textwidth}
        \centering
        \caption{Public Transport Class Prediction F1 Scores}
        \begin{tabular}{l c c c c}
            \hline\hline
            \multicolumn{1}{c}{Method} & \multicolumn{1}{c}{Random Forest} & \multicolumn{1}{c}{XGBoost} & \multicolumn{1}{c}{FCNN} & \multicolumn{1}{c}{\begin{tabular}[c]{@{}c@{}}Hierarchical \\ Clustering\end{tabular}} \\ \hline
            Original & \textbf{0.79} & \textbf{0.79} & 0.72 & 0.78\\
            ROS & 0.75 & 0.73 & 0.70 & 0.68 \\
            RUS & 0.69 & 0.69 & 0.65 & 0.68 \\
            SMOTE-NC & 0.78 & 0.74 & 0.59 & 0.78 \\
            ADASYN & 0.78 & \textbf{0.79} & 0.60 & 0.78 \\
            Borderline-SMOTE & 0.78 & 0.75 & 0.78 & 0.78 \\
            SVM-SMOTE & 0.78 & 0.75 & 0.75 & 0.77 \\
            AC-SMOTE & 0.78 & \textbf{0.79} & 0.67 & 0.78 \\
            \hline\hline
        \end{tabular}
    \end{minipage}%
    \hfill
    \begin{minipage}[b]{0.48\textwidth}
        \centering
        \caption{Driving Class Prediction F1 Scores}
        \begin{tabular}{l c c c c}
            \hline\hline
            \multicolumn{1}{c}{Method} & \multicolumn{1}{c}{Random Forest} & \multicolumn{1}{c}{XGBoost} & \multicolumn{1}{c}{FCNN} & \multicolumn{1}{c}{\begin{tabular}[c]{@{}c@{}}Hierarchical \\ Clustering\end{tabular}} \\ \hline
            Original & 0.82 & \textbf{0.83} & 0.73 & 0.82 \\
            ROS & 0.78 & 0.77 & 0.65 & 0.72 \\
            RUS & 0.72 & 0.37 & 0.11 & 0.72 \\
            SMOTE-NC & 0.81 & 0.79 & 0.71 & 0.82 \\
            ADASYN & 0.82 & \textbf{0.83} & 0.59 & 0.82 \\
            Borderline-SMOTE & 0.82 & 0.79 & 0.81 & 0.82 \\
            SVM-SMOTE & 0.82 & 0.79 & 0.80 & 0.80 \\
            AC-SMOTE & 0.81 & \textbf{0.83} & 0.72 & 0.82 \\
            \hline\hline
        \end{tabular}
    \end{minipage}
    
\end{table*}

Notably, the F1 score of the FCNN is the lowest in most combinations, especially for the Cycling class. It was also observed during its grid search optimization that its performance decreased with increasing number of layers beyond two layers, which is an indication of overfitting on the major class. 

The receiver operating characteristic (ROC) - area under curve (AUC) \cite{obuchowski2005roc} of AC-SMOTE used in conjunction with random forest clustering, which is the best-performing combination for the Cycling class, is calculated for Walking, Cycling, Public Transport, and Driving classes to be 0.94, 0.84, 0.91, and 0.91, respectively. This result aligns with previous comparisons, where the combined models are least accurate on the minority class. It is noteworthy that while the Walking class occupies only 17.8\% of the dataset, it is predicted with the highest AUC. 

The average precision (AP) \cite{buckland1994relationship} scores of the four classes are found from the area under the precision-recall curves of the AC-SMOTE and random forest model as 0.81, 0.27, 0.86, 0.88, respectively. A significantly lower AP can be observed from the Cycling class, which indicates that the minority class poses a significant challenge to the performance of the combined model. The difficulty in predicting Cycling samples may also result from overlapping features with other classes. The AP scores of other classes are in a similar range and in accordance to the amount of samples of each class in the dataset. This shows that samples of each class are given appropriate weight by AC-SMOTE such that the majority classes do not dominate the decision-making of the model. 

\section{Conclusion and Future Work}
Urban population data, such as transportation mode choice, often contain various degrees of class imbalance, which poses a significant challenge for data-driven models to correctly identify under-represented samples in the datasets. Although such models may demonstrate high overall accuracy, it is also important to evaluate their performance on minor classes to improve the efficiency of transportation resource planning and allocation. To this end, this study proposes an adaptive oversampling framework that creates synthetic data based on the overall distribution of the minor classes rather than depending on class-specific labels. By applying DBSCAN, the samples belonging to the minor classes are clustered based on their density, which allows more synthetic data to be created for sparse data points. Random forest and XGBoost models that are trained with data augmented by AC-SMOTE, while showing slight improvements on the major classes, also exhibit significantly higher F1 scores on minor classes. This combined approach is also validated on data from another city to ensure its generalizability and robustness.

Although AC-SMOTE displays noticeable improvement on minority travel mode prediction over other resampling methods, significant improvement is still required as its F1 score remains low. In future works, more advanced clustering methods are to be included and used in conjunction with AC-SMOTE to yield better performance. Moreover, feature engineering is to be investigated to enhance information extraction and augment clustering algorithms. In recent years, newer modes of mobility such as electric scooters and hybrid vehicles have received more widespread usage across the globe. Hence, newer datasets are required to be investigated to ensure the applicability of AC-SMOTE on real-world scenarios.

\end{document}